\documentclass[conference]{IEEEtran}
\IEEEoverridecommandlockouts
\usepackage{amsmath,amssymb,amsfonts}
\usepackage{algorithmic}
\usepackage{graphicx}
\usepackage{textcomp}
\usepackage{xcolor}
\usepackage{caption}
\usepackage{subcaption}
\usepackage{times}
\usepackage{soul}
\usepackage{graphicx}
\usepackage{hhline}
\usepackage{array}
\usepackage{float}
\usepackage{booktabs}
\usepackage{relsize}
\usepackage{balance}

\usepackage{biblatex}
\def\BibTeX{{\rm B\kern-.05em{\sc i\kern-.025em b}\kern-.08em
    T\kern-.1667em\lower.7ex\hbox{E}\kern-.125emX}}
\begin{document}

\title{Graph Mining under Data scarcity\\

}

\makeatletter
\newcommand{\linebreakand}{%
  \end{@IEEEauthorhalign}
  \hfill\mbox{}\par
  \mbox{}\hfill\begin{@IEEEauthorhalign}
}

\author{

\IEEEauthorblockN{Appan Rakaraddi}
\IEEEauthorblockA{
\textit{Nanyang Technological University}\\
Singapore \\
appan001@e.ntu.edu.sg}
\and
\IEEEauthorblockN{Lam Siew-Kei}
\IEEEauthorblockA{
\textit{Nanyang Technological University}\\
Singapore \\
assklam@ntu.edu.sg}
\and
\IEEEauthorblockN{Mahardhika Pratama}
\IEEEauthorblockA{
\textit{University of South Australia}\\
Australia \\
dhika.pratama@unisa.edu.au}
\and
\IEEEauthorblockN{Marcus de Carvalho}
\IEEEauthorblockA{
\textit{Nanyang Technological University}\\
Singapore \\
ivsucram@gmail.com}

}


\maketitle

\begin{abstract}

Multitude of deep learning models have been proposed for node classification in graphs. However, they tend to perform poorly under labeled-data scarcity. Although Few-shot learning for graphs has been introduced to overcome this problem, the existing models are not easily adaptable for generic graph learning frameworks like Graph Neural Networks (GNNs). Our work proposes an Uncertainty Estimator framework that can be applied on top of any generic GNN backbone network (which are typically designed for supervised/semi-supervised node classification) to improve the node classification performance. A neural network is used to model the Uncertainty Estimator as a probability distribution rather than probabilistic discrete scalar values. We train these models under the classic episodic learning paradigm in the $n$-way, $k$-shot fashion, in an end-to-end setting.

Our work demonstrates that implementation of the uncertainty estimator on a GNN backbone network improves the classification accuracy under Few-shot setting without any meta-learning specific architecture. We conduct experiments on multiple datasets under different Few-shot settings and different GNN-based backbone networks. Our method outperforms the baselines, which demonstrates the efficacy of the Uncertainty Estimator for Few-shot node classification on graphs with a GNN.

\end{abstract}


\section{Introduction}

Graphs are among the most ubiquitous form of data representation that are not constrained by the Euclidean geometry; and hence development of algorithms for graph mining is an active research area \cite{zhou2020graph}. Plethora of deep learning frameworks specifically crafted for graph-type topological structures, classed together as Graph Neural Networks (GNNs) \cite{gat,kim2022pure} have found applications in many domains like physics \cite{sanchez2018graph}, chemistry \cite{bradshaw2018generative}, Computer vision \cite{wang2019dynamic,qi2017pointnet}, Natural Language Processing \cite{bastings2017graph}, Transfer/Continual learning \cite{tang2021graphbased,rakaraddi2022reinforced}, and many other wide-range of areas \cite{wang2019zero,rakaraddi2021unsupervised}. But the most popular flavours of GNNs require copious amount of labeled training data for model generalization on unseen tasks, which raises a barrier for training the model on many real-world applications.

Scarcity of labeled training data for tuning the neural network models pioneered the research in Few-shot learning \cite{maml,proto,relation} to optimize the model with a relatively limited number of training data samples. Although these methods were originally proposed for image and language-based data classification, they have subsequently been extended to the graph domain with suitable adaptations. Three types of Few-shot learning on graphs include: node classification, edge classification and graph classification \cite{ijcai2022p789}. 
We focus specifically on node classification in this work. The existing works for node classification under Few-shot settings  \cite{meta-gnn,gmeta,rale,tent,gpn} use metric learning, optimization-based learning and hybrid architectures.

\textit{Metric learning} based methods \cite{gpn,gfl} map the raw data into an embedding space, and similarity measure such as the Euclidean distance or the Cosine Similarity compares the embedding distance between the support and query node embeddings for node classification.
\textit{Optimization-based learning} methods \cite{meta-gnn,rale} primarily leverage meta-learning approaches like MAML \cite{maml}, which is a model-agnostic task optimizer that can adapt to unseen tasks with the multi-level gradient descent  during the training. The hybrid architectures \cite{tent,gmeta} use a combination of the aforementioned methods to account for other graph topological information. Despite showing great performance under labeled-data scarcity conditions, these models use a specialised/specific architecture that cannot be easily adapted in combination with generic semi-supervised learning models. For example, let's consider a scenario where we are training a graph learning model for node classification across a series of $n$ tasks i.e., $\{\mathcal{T}_1,\mathcal{T}_2, ..., \mathcal{T}_n\}$ with all tasks having sufficiently labelled data, except for some task $\mathcal{T}_i \ (i \in [1,n])$. It may become computationally expensive and impractical to build a new model specifically for the task $\mathcal{T}_i$ under Few-shot settings, especially if the model has a lot of trainable parameters.

To mitigate this, our work proposes an additional training module called  \textbf{U}ncertainty on \textbf{G}raph \textbf{N}etworks (\textit{UGN}), that can be applied on top of any generic base model, which can be trained under Few-shot settings for the data-scarce task. This eliminates the need to develop an  extensive model for data-scarce conditions, which leads to a more resource-efficient model.
The data scarce task has a classification uncertainty arising with it and may lead to overfitting. Also, the common issue with overfitting is high variance values in the classification predictions between the training and the test sets. The UGN module which is applied on top of the base model estimates the class prediction uncertainty, which can effectively reduce overfitting problem. 

In summary, our work presents an uncertainty estimator framework called UGN for node classification on graphs under labeled training data scarcity, which can be applied on top of any generic graph embedding/GNN model to improve the classification accuracy. Although there are some existing works on uncertainty estimation \cite{zhang2020uncertainty,yuan}, to the best of our knowledge, our work is the first to extend this method to graphs for node classification.
Our contributions are listed as follows:
\begin{itemize}
    \item We propose an uncertainty estimator framework called \textit{UGN} for Few-shot node classification, that can be applied on top of any generic GNN for meta-learning.
    \item We conduct experiments on multiple datasets under different Few-shot settings as well as on different GNN variants to demonstrate the model adaptability and scalability.
    \item Our model outperforms the baseline methods, which demonstrates its effectiveness over the state-of-the-art.
    
\end{itemize}



\section{Related Work}

In this section, we review the literature that are relevant to our work.

\subsection{Graph Neural Networks}
The most common variant of graph learning methods are the Graph Neural Networks (GNNs), which map a graph's node and edge features into a low dimensional space, enabling a cost-efficient data mining. The most common form of GNNs are the message-passing GNNs (MP-GNNs) like GCN \cite{gcn}, GraphSAGE \cite{sage}, GAT \cite{gat}, SGC \cite{sgc}, GIN \cite{gin} and many other frameworks. These MP-GNNs differ from each other depending on the way the nodes and edges features are gathered and mixed together. 
Apart from the MP-GNNs, other methods have been proposed based on the WL-test \cite{wl}, which have proven to be more effective than the simple MP-GNNs.  This is so because MP-GNNs in certain cases have failed to distinguish 2 different graphs and have produced the same feature representations for both of them\cite{gin}. GNNs have also deployed  transformers-based framework \cite{transformer} to accommodate the graph heterogeneity \cite{gtn}. Also, the incorporation of  the positional/structural encoding \cite{lpse} in the GNNs has been shown to produce higher quality embedding.

\subsection{Few-shot Learning on Graphs}

Meta-GNN \cite{meta-gnn} proposes a method for Few-shot node classification using MAML \cite{maml} for graph-structured data. Graph Prototypical Networks (GPN) \cite{gpn} is another method proposed for Few-shot node classification with two components connected back-to-back: \textit{Network Encoder} and \textit{Network Valuator}. The Network Encoder generates the node representation embeddings using GNNs, while the Network Valuator estimates the node importance scores through a score aggregation layer. This entire network is optimized via Metric-learning similar to the Prototypical Networks \cite{proto}.
G-Meta \cite{gmeta} performs Few-shot node classification by  converting every node in the graph into a subgraph from the local neighbouring nodes. These embeddings of the subgraphs are used to generate the class prototypes and this model is optimized via MAML \cite{maml}.

\subsection{Uncertainty Learning}

Uncertainty-based learning has gained traction for dealing with scarce data-samples classification tasks. UAFS \cite{zhang2020uncertainty} proposed an uncertainty based method for estimating the classification uncertainty in Few-shot learning for image data. It utilizes a GNN to model the probability uncertainty as a Gaussian distribution. UCN \cite{yuan} proposes an uncertainty modeling of the classification outputs of the images using mutual information between the augmented query sample scores. The support and query sets are data augmented with $m$ different augmentation variants, and the query instance's classification probability are calculated using a metric-based approach for each of the augmentation. The uncertainty associated with the classification is modeled via Shannon-entropy function.

\section{Preliminaries}

Given a graph $\mathcal{G}(V,E,X,C)$, let $V \in \mathbb{R}^{|V|}$ represent the set of nodes, $E \in  \mathbb{R}^{|V|\times |V|}$ denote the edge set, $X \in \mathbb{R}^{|V| \times d}$ represent the node features matrix and $C \in \mathbb{R}^{|V|}$ represent the set of node classes. \newline

\textbf{Problem statement: }
The node classes $C = C_{train} \cup C_{novel}$ is split into $C_{train}$ (nodes seen during meta-training stage) and $C_{novel}$ (nodes seen during  meta-testing stage) such that $C_{train} \cap C_{novel}=\emptyset$. 
The objective of Few-shot node classification is to train the model on the sufficiently labeled $C_{train}$ classes and adapting it to classify the sparsely labeled $C_{novel}$ classes. Episodic learning is a popular paradigm employed for this training, where the model training is transformed as a series of tasks called episodes.
In each episode, the classes are sampled from the $C_{train}$ and $C_{novel}$ sets during the meta-training and meta-testing stages respectively. Each episode $i$ is split into \textit{support set} $\mathcal{S}$ and \textit{query set} $\mathcal{Q}$. The support set $\mathcal{S}$ consists of $n$ classes with $k$ node samples from each class, and the query set $\mathcal{Q}$ consists of the same $n$ classes with $m$ node samples from each of the $n$ classes. This is called \textit{n}-way \textit{k}-shot learning.

Let $\mathcal{S}_j$ represent the set of nodes belonging to the class $j$ in the support set $\mathcal{S}$. Then given a graph embedding network (e.g., a GNN) $f_{\theta}(.)$ with parameters $\theta$, the class prototype embedding of class $j$ is represented as:


\begin{equation}
    c_j = \frac{1}{|\mathcal{S}_j|} \sum_{u \in \mathcal{S}_j} f_{\theta}(u)
\end{equation}

where $|\mathcal{S}_j|$ depicts the number of support node samples belonging to class $j$. So in Metric-based learning, the probability of a query node $x$ belonging to the class $j$ is defined as:

\begin{equation}
    \label{eq:dist}
    p(j|x) = \frac{ \exp \big(-d(f_{\theta}(x),c_j) \big) }
    {\sum_i^n  \exp \big(-d(f_{\theta}(x),c_i) \big) }
\end{equation}

where $d(.)$ is a distance metric function like the Euclidean distance measure or the Cosine distance measure. The distance function is also often substituted with the similarity function $sim(.)$ (such as Cosine similarity) with appropriate sign change. This is written as:

\begin{equation}
    \label{eq:sim}
    p(j|x) = \frac{ \exp \big(sim(f_{\theta}(x),c_j) \big) }
    {\sum\limits_{i=1}^n  \exp \big(sim(f_{\theta}(x),c_i) \big) }
\end{equation} \newline

In our work, we use the Cosine similarity for the similarity function $sim(.)$. 

\section{Model Architecture}

\begin{figure*}[ht]
    \centering
    \includegraphics[scale=0.2]{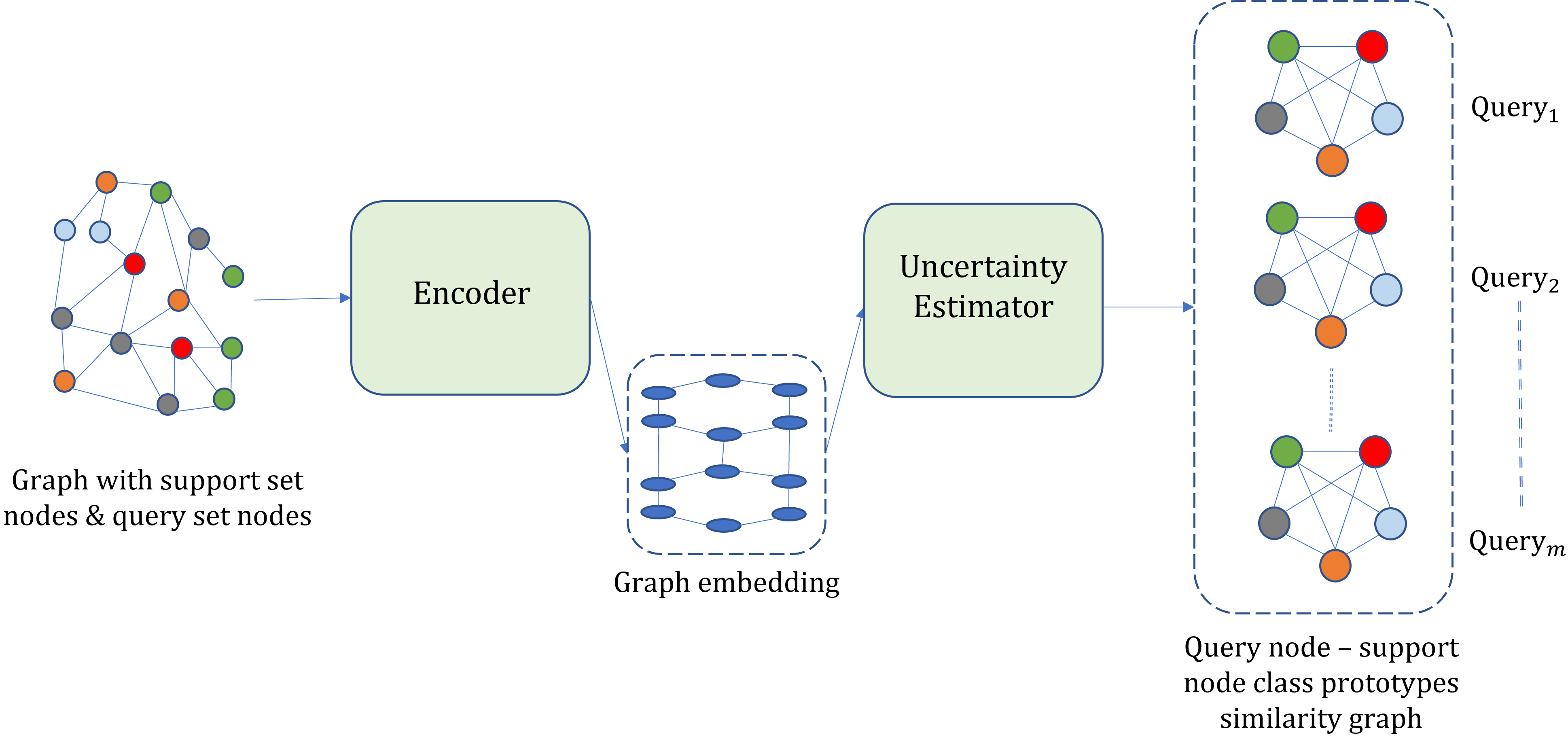}
    \caption{Model architecture of UGN: The graph is fed into the encoder, which is a generic MP-GNN model to generate the node embedding. The embedding is fed into the UGN layer to generate the class prototype similarity graph for each of the $m$ query samples. The deployment of the UGN layer to generate a probability distribution instead of training the model based on just on the Encoder probability outputs increases the average classification accuracy under labeled data scarcity conditions.}
    \label{fig:trainer}
\end{figure*}

Our model \textit{UGN} comprises of two cascaded components: \textit{Encoder} and \textit{Uncertainty Estimator}.

\subsection{Encoder} In our work, we have used a MP-GNN to embed the node features. We designed the Encoder as the backbone network with different variants of GNNs. 
For a given node $u$, the MP-GNN iteratively aggregates features of its neighbouring nodes, given by $\mathcal{N}(u)$,  and combines them together to generate the node $u$'s new embedded feature representation in the low-dimensional space. This is formally expressed as:

\begin{equation*}
\resizebox{0.47\textwidth}{!}{$
  \textbf{h}_u^{(k)} =   \text{{{\footnotesize UPDATE}}}\left( \textbf{h}_u^{(k-1)}, 
    \text{\footnotesize {AGGREGATE}} \left( \{ \textbf{h}_v^{(k-1)}, \  \forall v \in   \mathcal{N}(u)\} \right)  \right)$}
\end{equation*}

where $\textbf{h}_u^{(k)}$ represents the node embedding representation after $k$ iterations,  \text{\footnotesize {AGGREGATE}} and \text{{{\footnotesize UPDATE}}} represent the functions which gather the neighbouring node features and integrate into $u$ respectively. We represent the Encoder function as $f_{\theta}(.)$ and the embedding of a node $u$ as $f_{\theta}(u)$. 

As mentioned earlier, these embedding methods are designed for supervised learning settings in the presence of sufficient amount of data and generally perform poorly under labeled-data scarcity conditions. The Uncertainty Estimator which is added on top of the Encoder is discussed in the following section.

\subsection{Uncertainty Estimator} 
Our framework adopts a similar direction as UAFS \cite{zhang2020uncertainty}, which was designed for meta-learning with uncertainty estimator on image data.

To counteract the classification uncertainty in the model predictions due to data scarcity, the output of the Encoder which maps the nodes to their respective classes, is represented as probability distribution instead of probabilistic scalar values. This is achieved  by coupling the output of the Encoder with the input of the Uncertainty Estimator module.
We model the likelihood of a query node $x$ belonging to a class $j$  as a Gaussian distribution. This assumption is valid since the class-query pairs with the highest similarity values will have the highest probability value and the lower similarity values will be spread apart as deviations from the mean value.

Hence, the  probability  of the query node $x$ belonging to the class $j$ represented in the form of Gaussian distribution is formulated as:

\begin{equation}
\label{eq:normal}
    p(j|x) \sim \mathcal{N}(\mu_{(x,c_j)},\,\sigma_{(x,c_j)})\
\end{equation}

where $\sigma_{(x,c_j)}$ is the Standard Deviation that depicts the uncertainty in the classification prediction of the query node $x$ to the class $j$. The mean value $\mu_{(x,c_j)}$ is the fixed scalar similarity value between the query node embedding and the class prototype $c_j$, expressed as:
$$\mu_{(x,c_j)} = sim( f_{\theta} (x) , c_j )$$

The analytic expression represented in Equation \ref{eq:normal} is hard to estimate practically. So, we consider multiple discrete samples to approximate the Gaussian distribution via Monte-Carlo methods
that can easily be modeled by a neural network $g_{\phi}(.)$. Each sample represents the similarity value between a query embedding $f_{\theta}(x)$ and the class prototype $c_j$, with $T$ number of samples obtained for each of the pairs. The $sim_{(x,c_j)}$ associated with the $t$-th sample is given by:

\begin{equation}
    sim_{(x,c_j,t)} = \mu_{(x,c_j)} + \sigma_{(x,c_j)}\epsilon_t, \quad \epsilon_t \in \mathcal{N}(0,1)
\end{equation}

\subsection{Deviation Determination}


Since the probabilistic uncertainty of a query $x$ belonging to a class $j$ also depends on the probabilistic uncertainty of $x$ belonging to class $i$ (as the uncertainty associated with the classification of a node may overlap with another node's uncertainty), it is convenient to model the interdependence of the uncertainties as a graph.
The uncertainty of $x$ belonging to a class $j$, i.e., $\sigma_{(x,c_j)}$, is calculated between a query and each class prototype, where a node $j_x$ represents the Standard Deviation between the query sample $x$ and the class prototype $c_j$. To exploit this graph type structure, the neural network $g_{\phi}(.)$ is modeled using a GNN (in our case, we have used a GCN).
We split the  query embedding $f_{\theta}(x)$ and class prototype into $L$ equal parts. The feature vector of a class prototype node $j_x$ is constructed by taking the inner dot product between each of the $L$ pairs of the query embedding $f_{\theta}(x)$ and the class prototype $c_j \in [1,n]$ to generate the relational-similarity vector $\textbf{F} \in R^{n\times L}$.
The edges of the graph determine the similarity between two class prototype nodes. Following the earlier works  \cite{wang2018videos,zhang2020uncertainty}, we construct an edge between each pair of class prototype nodes $i_x$ and $j_x$ as:

\begin{equation}
    E(i_x,j_x) = \phi '(\textbf{F})^T \phi (\textbf{F}) 
\end{equation}

where $\phi(.)$ and $\phi '(.)$ are linear neural networks. The constructed graph is passed through a 2-layer GCN $g_{\phi}(.)$ to map the query sample $x$ and the class prototype pairings to their corresponding Standard Deviation values.

\subsection{Effective Similarity}

To calculate the effective similarity between a query and node prototype pairing, the mean of the normalized similarity scores is calculated. This is represented as:

\begin{equation}
\label{eq:sd}
    \textbf{sim}_{(x,c_j)} = \mathlarger{\frac{1}{T} {\sum_{t=1}^{T} }} \left(
    \frac{ \exp \big(sim_{(x,c_j,t)} \big) }
    {\mathlarger{\sum_i^n}  \exp \big(sim_{(x,c_i,t)}) } \right)
\end{equation}

Equation \ref{eq:sd} is optimized via the Negative Log Loss function during joint training of the Encoder and the Uncertainty Estimator. The entire model consisting of the \textit{Encoder} and \textit{UGN} is jointly  trained in an end-to-end manner as shown in Figure \ref{fig:trainer}.

 \section{Experiments}


 We have compared our model against GCN \cite{gcn}, GraphSAGE \cite{sage}, GAT \cite{gat}, GIN \cite{gin}, SGC \cite{sgc} and APPNP \cite{appnp} as base models and implemented the uncertainty layer UGN on top of these methods. All of these are trained and tested under the $n-$way $k-$shot paradigm. As mentioned in the introductory section, we have not compared our methods to the Few-shot specific architectures since the advantage of using our method lies in its adaptability with any generic base GNN architecture (that can be a part of other tasks or be a standalone task), unlike the Few-shot methods that rely on more specialised frameworks.


\begin{table}[ht]
\centering
\caption{Dataset properties and task-splitting}
\label{table:dataset}
\resizebox{\linewidth}{!}{%
\begin{tabular}{l||ccccc} 
\hhline{=:t:=====}
\textbf{Dataset} & \multicolumn{1}{l}{\textbf{Nodes \#}} & \multicolumn{1}{l}{\textbf{Edges \#}} & \multicolumn{1}{l}{\textbf{Features \#}} & \multicolumn{1}{l}{\textbf{Classes \#}} & \begin{tabular}[c]{@{}c@{}}\textbf{\textbf{Train/Val}}\\\textbf{\textbf{/Test}}\end{tabular} \\ 
\hhline{=::=====}
Amazon clothing & 24,919 & 91,680 & 9,034 & 77 & 40/17/20 \\
Amazon-Electronics & 42,318 & 43,556 & 8,669 & 167 & 90/37/40 \\
DBLP & 40,672 & 288,270 & 7,202 & 137 & 80/27/30 \\
\hline
\end{tabular}
}
\end{table}

We used 3 different datasets for comparison which are described below:

\begin{table*}
\centering
\caption{Average Accuracy values (in \%) of baseline methods on different datasets. The bold values indicate the highest accuracy values in a column.}
\label{table:accuracy}
\resizebox{\linewidth}{!}{%
\begin{tabular}{|c|cc|cc|cc|} 
\hline
~ & \multicolumn{2}{c|}{Amazon electronics ~ ~ ~} & \multicolumn{2}{c|}{DBLP ~ ~ ~} & \multicolumn{2}{c|}{Amazon clothing ~ ~} \\ 
\cline{2-7}
~ & 5-way 3-shot & 5-way 5-shot & 5-way 3-shot & 5-way 5-shot & 5-way 3-shot & 5-way 5-shot \\ 
\hhline{|=======|}
GCN & 36.7 & 25.4 & 35.0 & 23.9 & 35.3 & 27.1 \\
GraphSAGE & 29.9 & 23.0 & 36.2 & 28.1 & 38.7 & 28.7 \\
GAT & 32.3 & 27.0 & 57.6 & 25.5 & \textbf{58.3} & 27.2 \\
GIN & 26.6 & 22.2 & 20.0 & 21.8 & 28.6 & 25.4 \\
SGC & 38.6 & 26.5 & 41.5 & 27.3 & 46.9 & 31.0 \\
APPNP & 40.6 & 30.6 & 50.4 & 44.0 & 47.2 & 43.7 \\ 
\hline
UGN-GCN & \textbf{49.1} & \textbf{51.5} & 49.2 & 52.0 & 54.8 & 47.0 \\
UGN-GraphSAGE & 46.6 & 49.7 & 54.0 & 53.8 & 47.9 & 46.7 \\
UGN-GAT & 49.0 & 47.9 & \textbf{58.4} & \textbf{59.3} & 38.3 & 41.6 \\
UGN-GIN & 35.4 & 24.7 & 25.6 & 21.4 & 25.9 & 20.7 \\
UGN-SGC & 43.4 & 43.9 & 50.6 & 42.4 & 53.7 & \textbf{47.6} \\
UGN-APPNP & 48.1 & 47.2 & 28.8 & 33.9 & 49.9 & 42.5 \\
\hline
\end{tabular}
}
\end{table*}

\begin{itemize}
    \item \textbf{Amazon clothing:} \cite{amazon-clothing} This dataset is derived from the  Amazon products subcategory of "Clothing, Shoes and Jewelry", where 
    the nodes represent the products, and 
    an edge exists between a pair of nodes $(u,v)$ if a user who views product $u$ also views product $v$. The product review is described by the product's node feature. The node classes in the dataset are split as 40/17/20 for the train/validation/test sets.

    \item \textbf{Amazon-Electronics:} \cite{amazon-clothing} This dataset is derived from the Amazon products subcategory "Electronics". Each node denotes the product and the product reviews are considered as the node feature. A link exists between a pair of nodes if the two products are bought together. The node classes of the dataset are split as 90/37/40 for train/val/test sets.

    \item \textbf{DBLP:} \cite{dblp} This dataset is a subset of the  heterogeneous citation network from the DBLP computer science bibliography  website. The nodes represent the papers, and the edges represent the citations between them. The node classes of the dataset are split as 80/27/30 for train/val/test sets.


    
\end{itemize}
These details are summarised in Table \ref{table:dataset}.

\subsection{Implementation}

Our code is implemented using PyTorch framework and the GNNs are implemented using PyTorch-Geometric library \cite{pyg}. Our work uses two-layer GNNs with the hidden layer size of 16 nodes. We train the model for an average of 1000 episodes on each dataset. We have sampled the distribution values on an average of 1000 times to get a more precise value. All the experiments are carried out on 8GB GeForce GTX 1080 GPU hardware.

\subsection{Results}

\begin{figure*}[!htbp]
     \centering
     \begin{subfigure}[b]{0.4\textwidth}
         \centering
         \includegraphics[width=\textwidth]{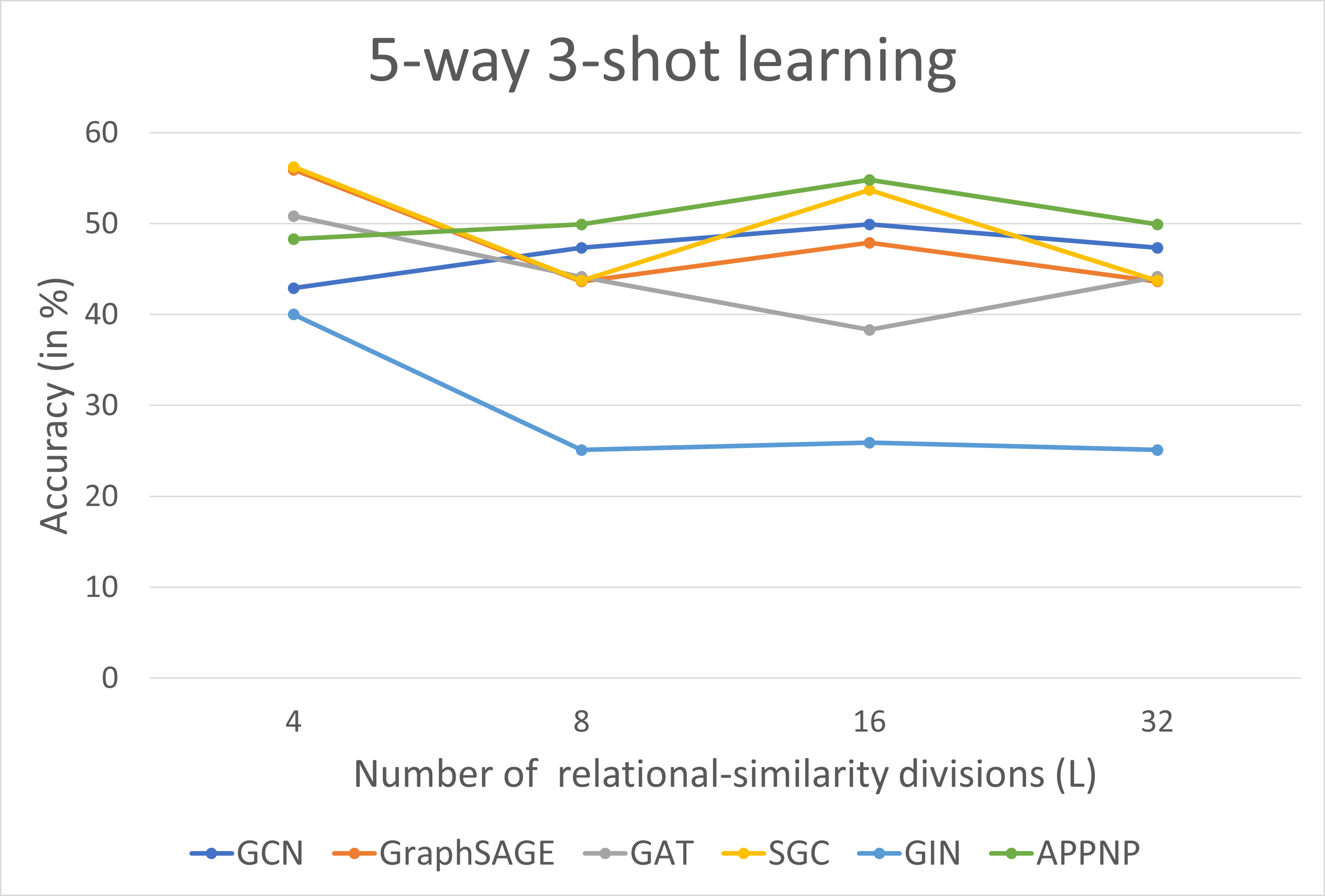}
         \label{fig:5-3}
     \end{subfigure}
     \hfill
     \begin{subfigure}[b]{0.4\textwidth}
         \centering
         \includegraphics[width=\textwidth]{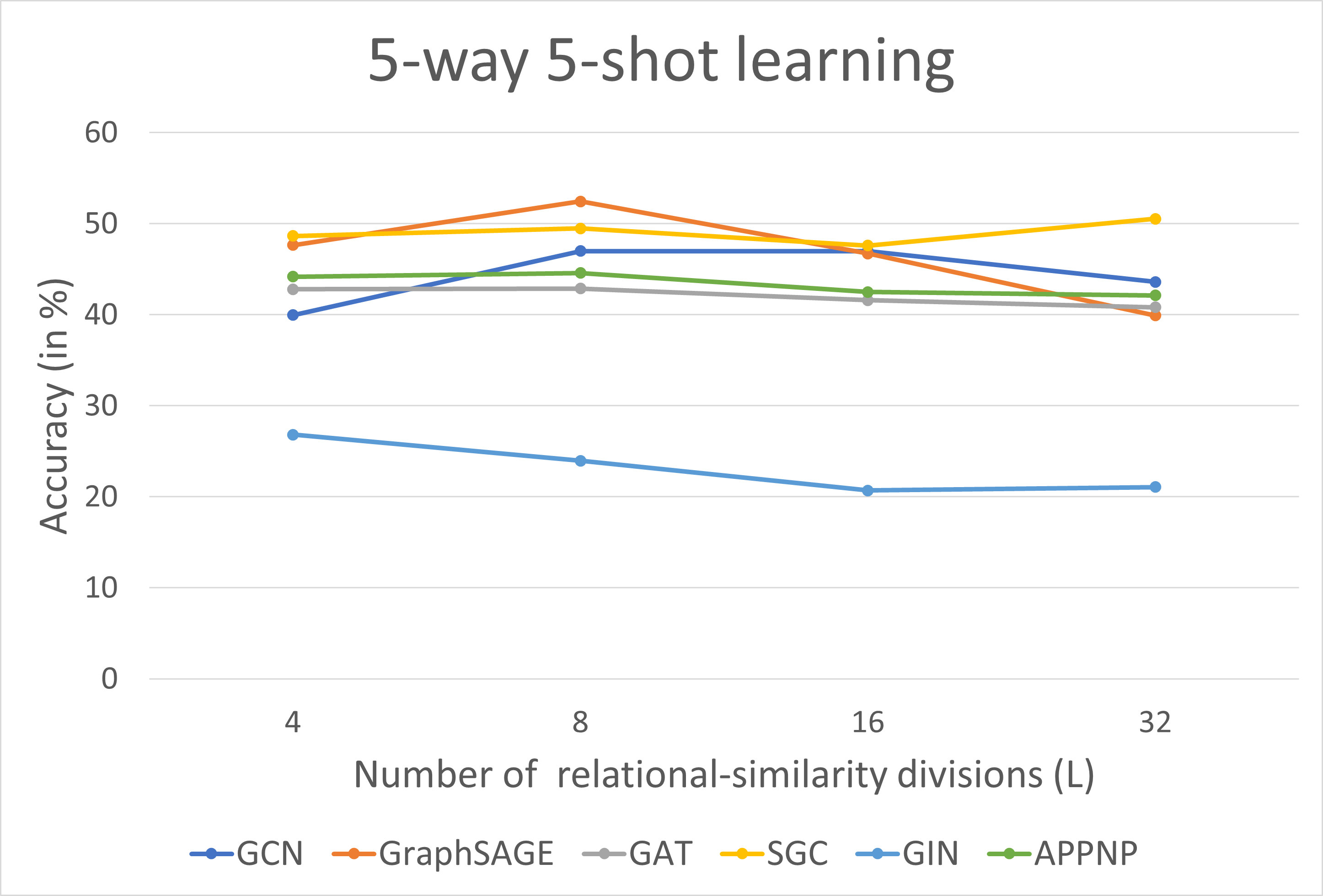}
         \label{fig:5-5}
     \end{subfigure}
        \caption{Comparison of average accuracy on the datasets under different Few-shot settings on the Amazon clothing dataset with varying number of partitions of the similarity vector.}
        \label{fig:graph-part}
\end{figure*}

We compare the results across the different baseline methods in Table \ref{table:accuracy}. The comparison metric used for evaluation is the average meta-test accuracy.
The performance on each of the datasets is compared under 2 different Few-shot settings: 5-way 3-shot learning and 5-way 5-shot learning. With the increasing number of training classes per episode, the average meta-test accuracy drops. It can be observed that across different base GNN models, the addition of the Uncertainty layer has consistently shown better performance in comparison to their counterparts, i.e., vanilla GNNs. Although we did observe a few outliers that violated the norm of improved accuracy under certain settings, our work in general has marked improvement in vast majority of the cases. For example, in the \textit{Amazon electronics} dataset, the addition of the UGN layer achieved a notable improvement in performance under both 5-way 3-shot and 5-way 5-shot settings for all different variants of GNN. For the dataset \textit{DBLP}, it showed a performance gain in all scenarios except UGN-APPNP and UGN-GIN (in 5-way 5-shot setting only).
The discrepancy in the result can be partially explained by the quality of the embedding generated by the MP-GNN and the aggregation scheme used by it, which negatively impacts the accuracy under low labeled-data conditions. But the accuracy in general across different datasets under different Few-shot settings has improved in comparison to the vanilla networks.

To demonstrate that the uncertainty layer UGN can perform effectively over any generic GNN, we also conducted experiments using different $g_{\phi}(.)$ in the UGN layer. Table \ref{table:accuracy-gat} shows the accuracy values on the \textit{Amazon clothing} dataset when using GAT as the $g_{\phi}(.)$ network in the UGN layer. The accuracy values have not significantly varied in comparison when using GCN for the UGN layer. This makes the UGN model agnostic of the core GNN i.e., the $g_{\phi}(.)$ network being used which demonstrates the universality of the method and the choice of the GNN making little to no effect on the output accuracy values.

\begin{table}
\centering
\caption{Average Accuracy values (in \%) of baseline methods on \textit{Amazon clothing} dataset with GAT as the base UGN. The bold values indicate the highest accuracy values in a column.}
\label{table:accuracy-gat}
\begin{tabular}{|c|cc|} 
\hline
~ & \multicolumn{2}{c|}{Amazon clothing ~ ~} \\ 
\cline{2-3}
~ & 5-way 3-shot & 5-way 5-shot \\ 
\hhline{|===|}
GCN & 35.3 & 27.1 \\
GraphSAGE & 38.7 & 28.7 \\
GAT & \textbf{58.3} & 27.2 \\
GIN & 28.6 & 25.4 \\
SGC & 46.9 & 31.0 \\
APPNP & 47.2 & 43.7 \\ 
\hline
UGN-GCN & 49.1 & \textbf{55.9} \\
UGN-GraphSAGE & 52.0 & 47.7 \\
UGN-GAT & 38.5 & 48.5 \\
UGN-GIN & 22.7 & 21.5 \\
UGN-SGC & 52.3 & 44.8 \\
UGN-APPNP & 48.3 & 43.0 \\
\hline
\end{tabular}
\end{table}

\subsection{Sensitivity Analysis}

To understand the effect of the number of partitions of the relational-similarity vector $\textbf{F} \in R^{n\times L}$ has on the accuracy values, we conducted multiple trials with different values of $L$ as shown in Figure \ref{fig:graph-part}. We partitioned the vector $\textbf{F}$ into 4, 8, 16 and 32 partitions and conducted the experiments with different GNN models on \textit{Amazon clothing} dataset under 5-way 3-shot and 5-way 5-shot settings. We observe in Figure \ref{fig:graph-part} that only GIN showed peak accuracy values with 4 partitions in both 5-way 3-shot and 5-way 5-shot settings, while the rest of them exhibit a varying behaviour. For example, GraphSAGE shows peak accuracy value for 5-way 3-shot learning at 4 partitions and at 8 partitions for 5-way 5-shot learning. A similar behaviour is observed for SGC, where the peak accuracy for 5-way 3-shot learning and 5-way 5-shot learning occur at different number of partitions. As mentioned earlier, the variation in the accuracy values for different MP-GNNs depending on the number of partitions is dependent on how the node neighbouring information is aggregated. So, the impact of the partitions for Uncertainty learning has a varying degree of effect on the meta-test accuracy dependent on base-GNN model used.


\section{Conclusion}
Our work introduces an Uncertainty Estimator network called UGN that can be applied on a generic GNN-backbone to achieve a significant performance gain under Few-shot learning. We tested our model on different variants of GNNs across multiple few-shot settings and different datasets to affirm the effectiveness of UGN. We also performed a sensitivity analysis to understand the effect of the number of partitions on the meta-test accuracy with different base GNN models. In conclusion, our model effectively demonstrates the effectiveness of UGN under Few-shot settings over a generic architecture.



\balance
\printbibliography

\end{document}